\documentclass[runningheads]{llncs}
\usepackage[T1]{fontenc}
\usepackage{amsmath}
\usepackage{float}
\usepackage{graphicx}
\usepackage{booktabs} 
\usepackage{multirow} 
\usepackage{siunitx}  
\usepackage{multirow}
\usepackage{subcaption}
\usepackage{siunitx}
\usepackage{amssymb}
\usepackage{tikz}
\usetikzlibrary{shapes.geometric, arrows.meta, positioning}

\begin{document}
\title{
Addressing Bias in VLMs for Glaucoma Detection Without Protected Attribute Supervision}
\titlerunning{Addressing Bias in VLMs for Glaucoma Detection}

%
%
\author{
  Ahsan Habib Akash\inst{1}
  \and
  Greg Murray\inst{1}
  \and
  Annahita Amireskandari\inst{1}
  \and
  Joel Palko\inst{1}
  \and 
  Carol Laxson\inst{1}
  \and 
  Binod Bhattarai\inst{2}
  \and
  Prashnna Gyawali\inst{1}
}
\authorrunning{AH Akash et al.}
\institute{
  \inst {}West Virginia University, Morgantown, WV, USA
  \and
  \inst {}University of Aberdeen, United Kingdom
}

\maketitle              
\begin{abstract}
Vision–Language Models (VLMs) have achieved remarkable success on multimodal tasks such as image–text retrieval and zero‐shot classification, yet they can demonstrate demographic biases even when explicit protected attributes are absent during training. In this work, we focus on automated glaucoma screening from retinal fundus images, a critical application given that glaucoma is a leading cause of irreversible blindness and disproportionately affects underserved populations. Building on a reweighting‐based contrastive learning framework, we introduce an attribute-agnostic debiasing method that (i) infers proxy subgroups via unsupervised clustering of image–image embeddings, (ii) computes gradient-similarity weights between the CLIP-style multimodal loss and a SimCLR‐style image‐pair contrastive loss, and (iii) applies these weights in a joint, top-$k$ weighted objective to upweight underperforming clusters. This label‑free approach adaptively targets the hardest examples, thereby reducing subgroup disparities. We evaluate our method on the Harvard–FairVLMed glaucoma subset, reporting Equalized-Odds Distance (EOD), Equal-
ized Subgroup AUC (ES-AUC), and Groupwise AUC to demonstrate equitable performance across inferred demographic subgroups.
\end{abstract}

\keywords{Unbiasing in VLMs  \and Retinal Diseases \and Glaucoma Classification \and Self-Supervised Learning}

\section{Introduction}

Glaucoma is a leading cause of irreversible blindness worldwide, with prevalence projected to rise in the coming years~\cite{quigley2006number}. Early detection is essential to prevent permanent vision loss, yet many patients remain undiagnosed due to the disease’s asymptomatic progression. This challenge is especially acute in rural and underserved regions, where access to specialist ophthalmic care is limited. 
In recent years, artificial intelligence, particularly deep learning models—has demonstrated significant promise in the automated diagnosis of diseases from medical imaging, including retinal fundus images. These models have achieved impressive accuracy in detecting conditions such as glaucoma~\cite{Saha2023,Fan2023,Rasel2024,Kruper2024,Chincholi2024}, streamlining workflows and enhancing early detection efforts.
Building on this progress, the emergence of vision-language models (VLMs) marks a notable advancement \cite{radford2021clip,li2023blip2}. By integrating both visual and textual inputs, VLMs can provide more comprehensive and context-aware interpretations, and are increasingly used in a variety of applications (e.g., image captioning, visual question answering).


In ophthalmology, VLMs can jointly interpret retinal fundus images and clinical text to not only detect glaucoma with high accuracy but also generate contextualized diagnostic reports (e.g., describing optic nerve head changes, cup‐to‐disc ratios, and recommended follow-up, that go beyond simple binary classification). By providing rich, narrative explanations alongside predictions, VLMs support more informed clinical decision-making and help bridge the gap in care, especially in low-resource settings.

As VLMs become integral to decision‐making pipelines, ensuring that they operate fairly across different demographic groups—particularly with respect to gender, race, and ethnicity—has become critically important. Unfortunately, large-scale pretraining data for foundation models are often imbalanced; prior work has documented systematic biases in VLMs that can lead to disparate treatment in settings ranging from criminal justice to healthcare \cite{hardt2016equality,chouldechova2017fair,ntoutsi2020bias}. For instance, a biased VLM might unfairly downweight clinical findings for underrepresented patient groups, potentially contributing to unequal diagnostic outcomes.

Most existing debiasing strategies depend on knowing sensitive attributes during training or fine‐tuning (e.g., explicitly using gender or race labels) \cite{ghanbarzadeh2023gender,xie2023parameter,xue2024bmft,jung2024unified,xue2025dfl}.
 In practice, however, collecting or storing such information is frequently infeasible due to legal, ethical, or regulatory constraints \cite{voigt2017gdpr}. Even when sensitive labels are available, many methods target only a single attribute at a time and do not guarantee fairness across other, implicitly encoded demographic factors. Furthermore, biases may be latent and not immediately apparent from the data distribution alone, so clustering or proxy‐based approaches that “guess” demographic groups can be unreliable: there is no clear way to verify whether the resulting clusters truly capture protected characteristics.

To this end, we propose learning unbiased representations for VLMs \emph{without} access to attribute labels.
Our approach builds on the principles of worst-case minimization for the image–text contrastive learning setting, optimizing for the lowest utility that any latent group might experience, thereby ensuring that no latent subgroup suffers disproportionately high loss—despite the absence of observed protected attributes.
We developed and evaluated our proposed framework using the Harvard-FairVLMED dataset, which provides detailed demographic attributes, ground-truth labels, and clinical notes to facilitate an in-depth examination of fairness within VL foundation models.

\begin{figure}[t]
  \centering
  \includegraphics[width=\linewidth]{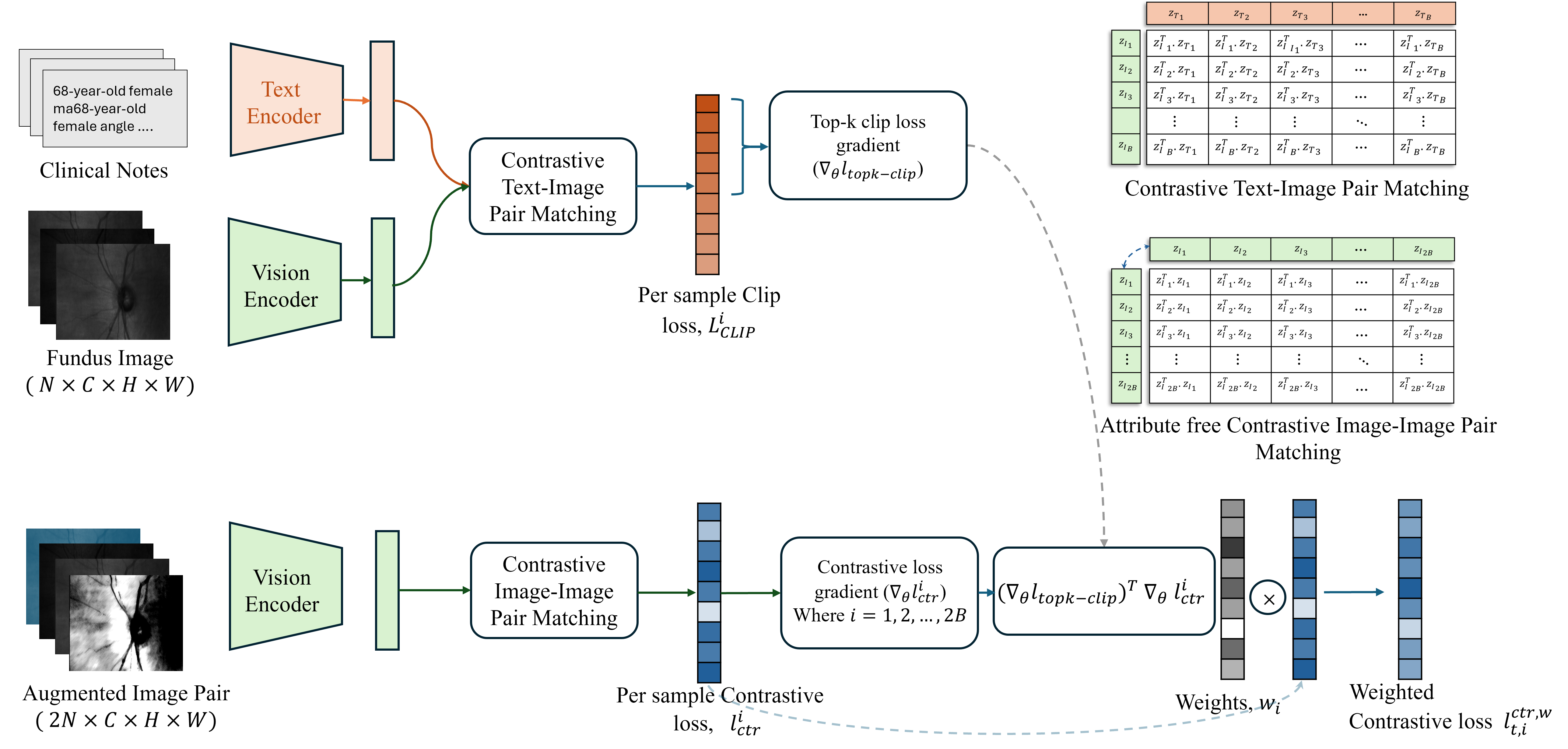}
  \caption{
    Overview of our \textbf{Debiased VLM} framework. We utilize a pretrained Vision-Language Model to align fundus images and clinical notes via the symmetric contrastive loss \(\mathcal{L}_{\mathrm{CLIP}}\). To enforce fairness without attribute labels, we apply a top-\(k\) CLIP loss to focus on the hardest image-text pairs. We compute gradient alignment between these top-\(k\) CLIP losses and per-sample contrastive image-pair losses, generating dynamic surrogate weights that reweight the contrastive loss to ensure the visual encoder learns unbiased representations while preserving multimodal alignment for glaucoma screening.}
  \label{fig:fairclip-surrogate}
\end{figure}

\section{Methodology}
\subsection{Problem Formulation}

Let \(\mathcal{D} = \{(x_i, t_i, y_i, a_i)\}_{i=1}^N\) be our training set, where, \(x_i\in \mathcal{X}\) is the \(i\)th fundus image, \(t_i\in \mathcal{T}\) is the \(i\)th clinical note, \(y_i\in \{0,1\}\) is the glaucoma diagnosis label, \(a_i\in \mathcal{A}\) is a protected attribute (e.g., race, gender, ethnicity). We process this multimodal data through a pretrained vision–language model (e.g.\ CLIP \cite{radford2021clip}), consisting of two encoders:
\begin{equation}
\label{eq:feature-extraction}
z_I^i = f_I(x_i; \theta_I), 
\quad
z_T^i = f_T(t_i; \theta_T),
\quad
f_I : \mathcal{X} \to \mathbb{R}^d,\;
f_T : \mathcal{T} \to \mathbb{R}^d.
\end{equation}
\noindent
Here, $f_I$ and $f_T$ are the image and text encoders that produce semantic and text features $z_I$ and $z_T$, where 
\(
z_I,\,z_T\in\mathbb{R}^d.
\)
For a minibatch of size $B$, define the per‐example losses
\[
\mathcal{L}_{\mathrm{I2T}}^{(i)}
= -\log\frac{\exp\bigl(s(z_I^i,z_T^i)/\tau\bigr)}
           {\sum_{j=1}^B\exp\bigl(s(z_I^i,z_T^j)/\tau\bigr)},\quad
\mathcal{L}_{\mathrm{T2I}}^{(i)}
= -\log\frac{\exp\bigl(s(z_I^i,z_T^i)/\tau\bigr)}
           {\sum_{j=1}^B\exp\bigl(s(z_I^j,z_T^i)/\tau\bigr)}.
\]
\noindent
where $s(u,v) = u^\top v$ and $\tau>0$ is a learnable temperature.  The symmetric CLIP loss over the batch is
\begin{equation}
\label{eq:clip-loss-latex}
\mathcal{L}_{\mathrm{CLIP}}(x_i,t_i;\theta_I,\theta_T,\tau)
= \frac{1}{2B}
\sum_{i=1}^B
\Bigl(
  \mathcal{L}_{\mathrm{I2T}}^{(i)} + \mathcal{L}_{\mathrm{T2I}}^{(i)}
\Bigr).
\end{equation}
\noindent
Therefore, the optimization goal can be defined as
\begin{equation}
\label{eq:optimization-latex}
\min_{\theta_I,\,\theta_T,\,\tau}
\;
\frac{1}{2B}
\sum_{i=1}^B
\Bigl[
  -\log\frac{\exp\!\bigl(s(z_I^i, z_T^i)/\tau\bigr)}
           {\sum_{j=1}^B \exp\!\bigl(s(z_I^i, z_T^j)/\tau\bigr)}
  \;-\;
  \log\frac{\exp\!\bigl(s(z_I^i, z_T^i)/\tau\bigr)}
           {\sum_{j=1}^B \exp\!\bigl(s(z_I^j, z_T^i)/\tau\bigr)}
\Bigr].
\end{equation}
In the context of debiasing, standard approach (e.g., FairCLIP \cite{Luo2024FairCLIP}) requires attribute information to be incorporated during model training. For visual encoder this can be described as $f_I : \mathcal{X} \times \mathcal{A} \rightarrow \mathbb{R}^d $. But such settings may not possible if attribute information is hidden from the training set. To ensure debiasing without explicit demographic data, Hashimoto et al. \cite{Hashimoto2018Fairness} enforced Rawlsian (worst-group) fairness by directly minimizing the maximum expected loss over all implicit groups. In order to guard against disparate performance across protected groups, Chai and Wang et al. \cite{chai2022self} treated the sensitive attribute as latent and enforced fairness by minimizing the worst-group contrastive representation error. For classification tasks, let $\mathcal{A}'$ denote the (unknown) set of all sensitive attribute values and, for each $a\in\mathcal{A}'$, define the index set $I_a = \{\,i : a_i = a\}$.  The fairness‐constrained learning problem is then cast as
\[
\min_{h}\;\max_{a\in\mathcal{A}'}
\;\frac{1}{|I_a|}\sum_{i\in I_a}
\mathcal{L}_{\mathrm{clf}}\bigl(h(x_i),y_i\bigr),
\]
where $\mathcal{L}_{\mathrm{clf}}(h(x_i),y_i)$ is the standard classification loss (e.g.\ cross‐entropy) for example $(x_i,y_i)$.  In this way, the learner finds a hypothesis $h$ that minimizes the highest average loss incurred by any (latent) sensitive group, thereby guaranteeing balanced performance across all subpopulations.

\subsection{Debiasing Visual Encoding with Surrogate Weighting}

Our objective is to train a visual encoder $f_I:\mathcal{X}\to\mathbb{R}^d$ whose representations minimize disparities among the hardest modeled examples, yet without access to protected attributes. For a minibatch of $N$ images, two stochastic augmentations $\{\tilde x_i^{(1)},\tilde x_i^{(2)}\}_{i=1}^N$ produce $2B$ views.  The standard image-pair contrastive loss
\begin{equation}\label{eq:ctr-loss}
\ell_{\mathrm{ctr}}(\tilde x_i,\tilde x_j)
= -\log\frac{\exp(s(f_I(\tilde x_i),f_I(\tilde x_j))/\tau)}{\sum_{m\neq i}\exp(s(f_I(\tilde x_i),f_I(\tilde x_m))/\tau)},
\quad s(u,v)=u^\top 
\end{equation} 
\noindent
But this will weight each pair equally, which unintentionally dilutes the effect of those views that incur the highest loss (i.e., the "hardest" views).  To enforce fairness across these difficult cases, we define the top-$k$ losses

\begin{equation}\label{eq:topk-ctr}
\ell_{\mathrm{topk\text{-}ctr}}
= \frac{1}{k}\sum_{m=1}^{2B}[\ell_{\mathrm{ctr}}(\tilde x_m)-\lambda_{\mathrm{ctr}}]_+ + \lambda_{\mathrm{ctr}}.
\end{equation}
where $\lambda_{\mathrm{ctr}}$ is the $k$-th largest loss. This relaxation acts as a smooth surrogate for the original Min–Max fairness objective, allowing us to focus more strongly on the hardest examples during backpropagation. In standard unsupervised contrastive learning, however, every non‐positive sample is treated as a negative, which invariably introduces false negatives pairs that actually share the same semantic or demographic characteristics but are penalized nonetheless. These false negatives warp the loss landscape and prevent the model from learning equally representative features for minority groups. To address this, we switch to the top-$k$ average CLIP loss as our primary training objective. By isolating the $k$ hardest image–text pairs, we capture the most informative gradient signals and then use those gradients to dynamically reweight the contrastive loss for each image pair. This ensures that bias-inducing errors receive proportionally greater corrective emphasis, yielding a fairer visual encoder. We can formulate the top-$k$ CLIP loss as

\begin{equation}\label{eq:topk-clip}
\ell_{\mathrm{topk\text{-}clip}}
= \frac{1}{k}\sum_{m=1}^{2B}[\mathcal{L}_{\mathrm{CLIP}}(\tilde x_m,t_m)-\lambda_{\mathrm{clip}}]_+ + \lambda_{\mathrm{clip}}.
\end{equation}
\noindent
where $ \lambda_{\mathrm{clip}}$ is the $k$-th largest loss. Now we want to focus only on those pairs whose corrective update aligns with reducing CLIP’s top-$k$ loss, so we measure gradient alignment for each view $m$:
\[
w_m \;=\;
\bigl(\nabla_{\theta_I}\,\ell_{\mathrm{topk\!-\!clip}}\bigr)^\top
\nabla_{\theta_I}\,\ell_{\mathrm{ctr}}(\tilde x_{m}),
\qquad
\hat w_m \;=\;\max\{w_m,0\}.
\]
normalizing to form a distribution:
\begin{equation}\label{eq:weights}
W_m=\frac{\hat w_m}{\sum_{j=1}^{2N}\hat w_j + \delta(\sum_{j=1}^{2N}\hat w_j)},
\quad\delta(r)=\begin{cases}1,&r=0,\\0,&\text{otherwise,}\end{cases}
\end{equation}
\noindent
This procedure yields a set of normalized weights $\{W_m\}_{m=1}^{2N}$ that dynamically emphasize only those contrastive pairs whose updates most directly reduce the top-$k$ CLIP loss.  We refer to this overall framework as \textbf{Debiased VLM}.  In Debiased VLM, each visual sample’s contrastive loss is reweighted according to its alignment with the hardest CLIP‐based gradient signals, ensuring that the minority subpopulations receive the highest weight.  A schematic of the full training loop is presented in Figure~\ref{fig:fairclip-surrogate}. 
Table \ref{tab:method-comparison} compares our objective with the two baselines used in our experiments. The standard CLIP~\cite{radford2021clip} model optimizes only the contrastive loss $\mathcal{L}_{\mathrm{CLIP}}$, while FairCLIP\cite{Luo2024FairCLIP} further incorporates a supervised adversarial term that requires explicit protected‐attribute labels. In contrast, our Debiased CLIP method applies an unsupervised surrogate weighting mechanism—based on per‐sample image–image gradient norms—to adaptively upweight hard subgroups without any demographic supervision.

\begin{table}[t]
  \centering
  \caption{Comparison of Debiasing Methods}
  \label{tab:method-comparison}
  \begin{tabular}{lcc}
    \toprule
    \textbf{Method} & \textbf{Required Supervision} & \textbf{Objective} \\
    \midrule
    CLIP  
      & None  
      & $\displaystyle \mathcal{L}_{\mathrm{CLIP}}$ \\[6pt]
    FairCLIP  
      & Protected attribute   
      & $\displaystyle \mathcal{L}_{\mathrm{CLIP}} \;+\;\lambda\,\mathbb{E}_{(I,T)}\bigl[D_{\mathrm{adv}}\bigl(f(I),\,g(T),\,a\bigr)\bigr]$ \\[10pt]
    Debiased CLIP  
      & None  
      & $\displaystyle \sum_{i} w_{i}\,\mathcal{L}_{\mathrm{CLIP}}(I_{i},T_{i}),\quad
         w_{i}\propto \bigl\lVert\nabla_{\theta}\mathcal{L}_{II}(I_{i};\theta)\bigr\rVert$ \\
    \bottomrule
  \end{tabular}
\end{table}

\section{Experiments}

\subsection{Dataset}
Our study makes use of the Harvard–FairVLMed dataset, which consists of 10\,000 de‐identified patient records drawn from 10\,000 unique individuals. We follow the original split, allocating 7\,000 records for training, 1\,000 for validation, and 2\,000 for testing. The mean age at the time of record collection is 60.9 years (SD = 16.2). In terms of race and ethnicity, 819 subjects identify as Asian, 1\,491 as Black, and 7\,690 as White; 90.6\% of participants are non‐Hispanic, 4.0\% are Hispanic, and 5.4\% have unspecified ethnicity. Female subjects account for 56.3\% of the cohort, with males comprising the remainder. Regarding preferred language, 92.5\% of patients primarily use English, 1.7\% use Spanish, 0.8\% use other languages, and 5.0\% have no language specified. Marital status is recorded as follows: 57.4\% are married or partnered, 26.4\% are single, 6.6\% have experienced divorce, 1.0\% are legally separated, 6.1\% are widowed, and 1.5\% have unspecified status. Finally, after de‐identification, the clinical notes range from 11 to 332 words, with an average length of 147 words per note.
\subsection{Pre-Training, Evaluation and Performance Metrics}

For our VLM experiments, we load the official checkpoint from  the pre‐trained CLIP model \cite{radford2021clip}, and fine-tune it on the Harvard–FairVLMed dataset. The image encoder $f_{I}$ and a text encoder $f_{T}$ of the CLIP model map inputs into a shared $d$‐dimensional embedding space. To evaluate performance on the Harvard–FairVLMed dataset without any additional fine‐tuning on classification, we adopt a zero‐shot classification protocol. Specifically, let $\mathcal{C} = \{c_{1}, c_{2}, \dots, c_{M}\}$ denote the set of diagnostic categories or clinical labels of interest (e.g., “glaucoma,” “no glaucoma,” etc.). For each class $c_{j}$, we construct a natural‐language prompt of the form  (I) “This image contains glaucoma” (ii) "This image does not contain glaucoma". Each prompt is passed through the text encoder to produce a text embedding $z_{T} \;=\; 
f_{T}\bigl(\,\text{"This image contains/does not contain glaucoma"}\bigr)\ $. At test time, given a held‐out fundus image $x$, we compute its image embedding $z_{I} \;=\; f_{I}(x)$.
Then, we compute cosine similarities between the image embedding \(z_I\) and each class prompt embedding \(z_{T,j}\), and assign the class with the highest similarity:
\begin{equation}
\hat{y}
\;=\;
\arg\max_{j\in\{1,\dots,M\}}
\frac{z_I^\top z_{T,j}}{\|z_I\|\;\|z_{T,j}\|}.
\end{equation}
\noindent
In this zero‐shot setting, no CLIP parameters are updated on Harvard–FairVLMed; instead, we rely on CLIP’s pre‐trained alignment to generalize to medical imagery and associated clinical concepts. This approach enables direct evaluation of CLIP’s zero‐shot capabilities on fundus images and clinical notes from the Harvard–FairVLMed dataset.

\noindent
\textbf{Performance Metrics:} 
We evaluate the performance of Debiased CLIP using different metrics: 1) Equalized Odds Difference (EOD), which measures the sum of the largest differences in TPR and FPR between any two subgroups; 2) Equalized Subgroup AUC (ES‐AUC), which measures the average of subgroup AUCs, penalizing large disparities; and 3) Group‐Wise AUC, where AUCs are computed separately for each subgroup. All values are reported as percentages. This suite of metrics captures both global performance and equity across subpopulations.


\begin{table}[t]
  \centering
  \setlength{\tabcolsep}{5pt}
  \small

  \caption{Zero‐shot transfer results of CLIP vs.\ Debiased CLIP on Harvard-FairVLMed. We report Equalized-Odds Distance (EOD $\downarrow$), Equalized Subgroup AUC (ES-AUC $\uparrow$), and group-wise AUCs (\%) for each protected attribute.}
  \label{tab:zeroshot_updated}

  \resizebox{\textwidth}{!}{%
  \begin{tabular}{
    @{} l
    c
    c
    c
    c
    c
    c
    c
    c
    @{}
  }
    \toprule
\textbf{Attribute} & \textbf{Model} & {EOD$\downarrow$} & {ES-AUC$\uparrow$} & \multicolumn{5}{c}{\textbf{Group-wise AUC (\%)}} \\
\cmidrule(lr){5-9}

    \multirow{4}{*}{Race} 
    &                  &                  &                   & Asian & Black & White &  &  \\
    & $\text{CLIP}_{\text{ViT-L}}$      & \textbf{24.00} & 64.90 & \textbf{70.85} & 65.99 & \textbf{67.99} &  &  \\
    & $\text{Debiased CLIP}_{\text{ViT-L}}$  & 36.36 & \textbf{65.49} & 68.52 & \textbf{67.29} & 66.42 & &  \\

    \midrule

    \multirow{4}{*}{Gender} 
    &                  &                  &                   & Female & Male &  & &  \\
    & $\text{CLIP}_{\text{ViT-L}}$      & \textbf{5.23}  & \textbf{66.21} & \textbf{66.93} & 70.03 &  &  &  \\
    & $\text{Debiased CLIP}_{\text{ViT-L}}$  & 11.47 & 62.77 & 64.09 & \textbf{70.88} &  &  & \\

    \midrule

    \multirow{4}{*}{Ethnicity} 
    &                  &                  &                   & Non-Hispanic & Hispanic & & &  \\
    & $\text{CLIP}_{\text{ViT-L}}$      & 17.10 & 61.64 & \textbf{68.66} & 57.92 &  &  &  \\
    & $\text{Debiased CLIP}_{\text{ViT-L}}$  & \textbf{3.50}  & \textbf{64.40} & 67.18 & \textbf{63.09} &  &  &  \\

    \midrule

    \multirow{4}{*}{Language} 
    &                  &                  &                   & English & Spanish & Others & &  \\
    & $\text{CLIP}_{\text{ViT-L}}$      & 22.28 & 57.66 & \textbf{68.44} & 58.24 & 60.08 &  &  \\
    & $\text{Debiased CLIP}_{\text{ViT-L}}$  & \textbf{16.00} & \textbf{60.97} & 66.74 & \textbf{70.17} & \textbf{60.52} &  &  \\

    \midrule

    \multirow{4}{*}{Age Group} 
    &                  &                  &                   & Young & Old & &  &  \\
    & $\text{CLIP}_{\text{ViT-L}}$      & 27.72 & 58.26 & 53.79 & \textbf{70.95} & & &  \\
    & $\text{Debiased CLIP}_{\text{ViT-L}}$  & \textbf{16.24} & \textbf{58.89} & \textbf{54.59} & 68.42 &  & &  \\

    \midrule

    \multirow{4}{*}{Marital Status} 
    &                  &                  &                   & Married & Single & Divorced & Separated & Widowed \\
    & $\text{CLIP}_{\text{ViT-L}}$      & 46.88 & 40.43 & \textbf{52.90} & \textbf{70.45} & 62.08 & \textbf{80.63} & 55.61 \\
    & $\text{Debiased CLIP}_{\text{ViT-L}}$  & \textbf{28.57} & \textbf{49.33} & 52.46 & 68.80 & \textbf{62.15} & 72.25 & \textbf{66.75} \\

    \bottomrule
  \end{tabular}
  }
\end{table}

\subsection{Results}

We first evaluate the performance of Debiased CLIP in comparison to the standard CLIP, using Equalized Odds Difference (EOD), Equalized Sensitivity AUC (ES-AUC), and group-wise AUCs across protected attributes (Table \ref{tab:zeroshot_updated}). Across six different attributes, we find that Debiased CLIP outperforms CLIP on four attributes based on EOD and ES-AUC. Even in cases where CLIP achieves a better EOD, the performance of minority groups often improves. For instance, the minority Black subgroup (constituting 14.91\% of the cohort) sees an AUC increase from 65.99 to 67.29, indicating enhanced performance for this underrepresented group despite a slightly higher EOD.
For attributes where Debiased CLIP performs better, the improvements are often substantial. For example, for ethnicity, EOD drops significantly from 17.10 to 3.50, while ES-AUC rises from 61.64 to 64.40. The Hispanic subgroup (4.0\% of the dataset) improves in AUC from 57.92 to 63.09, reflecting meaningful gains for this small minority group. Similarly, for language, EOD decreases from 22.28 to 16.00 while ES-AUC increases 57.66 to 60.97. The Spanish-speaking subgroup (1.7\% of the dataset) sees a major AUC increase from 58.24 to 70.17, substantially reducing language-related disparities.
These consistent improvements for minority or underrepresented subgroups across attributes are directly reflected in ES-AUC gains. This demonstrates that our debiasing strategy enhances fairness by improving worst-group performance while maintaining strong overall discrimination. Importantly, this analysis highlights the nuanced nature of fairness evaluation in medical VLMs. Even in cases where EOD does not decrease (e.g., Race), improvements in minority subgroup performance and ES-AUC suggest progress toward more equitable model behavior achieved without requiring explicit demographic labels during training.

\subsection{DebiasedCLIP vs FairCLIP}

\begin{figure}[t]
    \centering
    \includegraphics[width=0.95\textwidth]{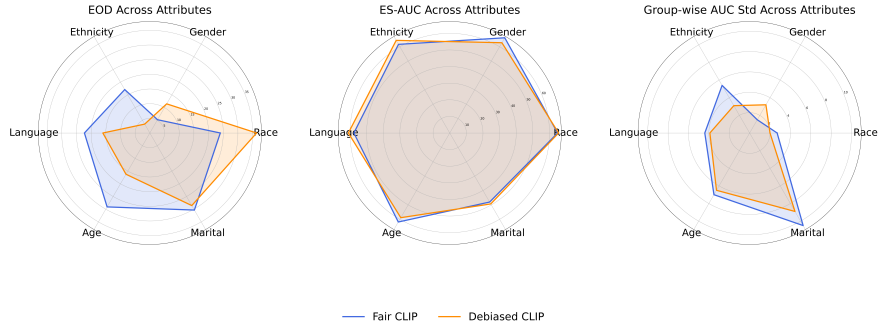}
    \caption{Comparison of Fair CLIP and Debiased CLIP using radar plots across Race, Gender, Ethnicity, and Language on the Harvard-FairVLMed dataset. Panels show Equalized Odds Distance (EOD, lower is better), Equalized Subgroup AUC (ES-AUC, higher is better), and group-wise AUC variability (lower indicates fairer consistency).}
    \label{fig:debiased_vs_fairclip}
\end{figure}

\begin{table}[t]
    \centering
    \caption{Group-wise AUC (\%) comparison between Fair CLIP and Debiased CLIP across demographic groups with dataset percentages on Harvard-FairVLMed.}
    \label{tab:group_auc_demographics}
    \vspace{8pt}
    \setlength{\tabcolsep}{8pt} 
    \resizebox{.85\textwidth}{!}{%
    \begin{tabular}{l l c c c}
        \toprule
        \textbf{Attribute} & \textbf{Group} & \textbf{\% in Data} & \textbf{FairCLIP} & \textbf{DebiasedCLIP} \\
        \midrule
        \multirow{3}{*}{Race} 
            & Asian & 8.2\% & 70.9 & 68.5 \\
            & Black & 14.9\% & 66.0 & 67.3 \\
            & White & 76.9\% & 68.0 & 66.4 \\
        \midrule
        \multirow{2}{*}{Gender} 
            & Female & 56.3\% & 66.9 & 64.1 \\
            & Male   & 43.7\% & 70.0 & 70.9 \\
        \midrule
        \multirow{2}{*}{Ethnicity} 
            & Non-Hispanic & 90.6\% & 68.7 & 67.2 \\
            & Hispanic     & 4.0\%  & 57.9 & 63.1 \\
        \midrule
        \multirow{3}{*}{Language} 
            & English & 92.5\% & 68.4 & 66.7 \\
            & Spanish & 1.7\%  & 58.2 & 70.2 \\
            & Others  & 0.8\%  & 60.1 & 60.5 \\
        \midrule
        \multirow{2}{*}{Age Group}
            & Young & 23.05\% & 58.86 & 54.59 \\
            & Old   & 76.95\% & 72.92 & 68.12 \\
        \midrule
        \multirow{5}{*}{Marital Status}
            & Married/Partnered & 57.4\% & 53.8 & 52.5 \\
            & Single            & 26.4\% & 69.6 & 68.8 \\
            & Divorced          & 6.6\%  & 61.5 & 62.2 \\
            & Separated         & 1.0\%  & 75.4 & 72.3 \\
            & Widowed           & 6.1\%  & 60.7 & 66.8 \\
        \bottomrule
    \end{tabular}
    }
\end{table}
We compare FairCLIP—which trains separate, attribute-specific models using sensitive labels—with our single-model, unsupervised Debiased CLIP. As Table \ref{tab:group_auc_demographics} and Figure \ref{fig:debiased_vs_fairclip} show, Debiased CLIP matches or outperforms FairCLIP on most metrics: it yields lower EOD for Ethnicity, Language, and Age, and equal-or-better ES-AUC for Race, Ethnicity, and Language, while also reducing AUC variability across subgroups. Although FairCLIP achieves slightly lower EOD on Race and Gender, Debiased CLIP boosts minority subgroup AUCs (e.g., Black in Race, Hispanic in Ethnicity). These findings demonstrate that our attribute-agnostic weighting delivers competitive fairness without demographic labels or multiple models, facilitating privacy-preserving clinical VLM deployment.

\section{Discussion}

Our results show that attribute-free surrogate weighting enhances fairness in medical VLMs without sensitive labels or multiple models. Within a single model, it preserves zero-shot glaucoma detection and raises ES-AUC for Race, Ethnicity, and Language—improving minority subgroup performance and reducing AUC variability. Although Fair CLIP attains lower EOD in some metrics, Debiased CLIP narrows clinically relevant disparities for smaller minority groups without compromising overall accuracy. This demonstrates that attribute-free debiasing is a practical, privacy-preserving approach for equitable VLM deployment—and could be extended to downstream tasks like progression risk prediction or treatment recommendation.

\section{Conclusion}
We introduce an attribute-agnostic debiasing framework for VLMs in glaucoma screening, using unsupervised clustering to define proxy subgroups and a surrogate weighting scheme that aligns difficult image–text gradients with per-sample image–image losses. This adaptively upweights underperforming clusters without any protected labels. On the Harvard–FairVLMed glaucoma subset, Debiased CLIP lowers Equalized-Odds Distance, raises Equalized Subgroup AUC, and preserves strong zero-shot discrimination. Compared to FairCLIP—which relies on demographic labels and separate models—our single-model approach delivers equal or better fairness across Race, Ethnicity, Language, and Age, offering a practical, privacy-preserving solution for clinical VLM deployment.

\section{Acknowledgment}
This work was supported by the Defense Advanced Research Projects Agency under Grant AI-CRAFT AWD16069, and by the National Institutes of Health under Grant P20 GM144230 from the National Institute of General Medical Sciences (NIGMS).

\bibliographystyle{IEEEtran}



\bibliography{main.bib}

\begin{thebibliography}{10}
\providecommand{\url}[1]{#1}
\csname url@samestyle\endcsname
\providecommand{\newblock}{\relax}
\providecommand{\bibinfo}[2]{#2}
\providecommand{\BIBentrySTDinterwordspacing}{\spaceskip=0pt\relax}
\providecommand{\BIBentryALTinterwordstretchfactor}{4}
\providecommand{\BIBentryALTinterwordspacing}{\spaceskip=\fontdimen2\font plus
\BIBentryALTinterwordstretchfactor\fontdimen3\font minus \fontdimen4\font\relax}
\providecommand{\BIBforeignlanguage}[2]{{%
\expandafter\ifx\csname l@#1\endcsname\relax
\typeout{** WARNING: IEEEtran.bst: No hyphenation pattern has been}%
\typeout{** loaded for the language `#1'. Using the pattern for}%
\typeout{** the default language instead.}%
\else
\language=\csname l@#1\endcsname
\fi
#2}}
\providecommand{\BIBdecl}{\relax}
\BIBdecl

\bibitem{quigley2006number}
H.~A. Quigley and A.~T. Broman, ``The number of people with glaucoma worldwide in 2010 and 2020,'' \emph{British Journal of Ophthalmology}, vol.~90, no.~3, pp. 262--267, 2006.

\bibitem{Saha2023}
S.~Saha, J.~Vignarajan, and S.~Frost, ``A fast and fully automated system for glaucoma detection using color fundus photographs,'' \emph{Scientific Reports}, vol.~13, p. 18408, 2023.

\bibitem{Fan2023}
R.~Fan, K.~Alipour, C.~Bowd, M.~Christopher, N.~Brye, J.~A. Proudfoot, M.~H. Goldbaum, A.~Belghith, C.~A. Girkin, M.~A. Fazio, J.~M. Liebmann, R.~N. Weinreb, M.~Pazzani, D.~Kriegman, and L.~M. Zangwill, ``Detecting glaucoma from fundus photographs using deep learning without convolutions: Transformer for improved generalization,'' \emph{Ophthalmology Science}, vol.~3, no.~1, p. 100233, 2023.

\bibitem{Rasel2024}
R.~K. Rasel, F.~Wu, M.~Chiariglione, S.~S. Choi, N.~Doble, X.~R. Gao, and \emph{et al.}, ``Assessing the efficacy of 2d and 3d cnn algorithms in oct-based glaucoma detection,'' \emph{Scientific Reports}, vol.~14, p. 11758, 2024.

\bibitem{Kruper2024}
J.~Kruper, A.~Richie-Halford, N.~C. Benson, S.~Caffarra, J.~Owen, Y.~Wu, and \emph{et al.}, ``Convolutional neural network–based classification of glaucoma using optic radiation tissue properties,'' \emph{Communications Medicine}, vol.~4, p.~72, 2024.

\bibitem{Chincholi2024}
F.~Chincholi and H.~Koestler, ``Transforming glaucoma diagnosis: transformers at the forefront,'' \emph{Frontiers in Artificial Intelligence}, vol.~7, p. 1324109, 2024.

\bibitem{radford2021clip}
A.~Radford, J.~W. Kim, C.~Hallacy, A.~Ramesh, G.~Goh, S.~Agarwal, G.~Sastry, A.~Askell, P.~Mishkin, J.~Clark, G.~Krueger, and I.~Sutskever, ``Learning transferable visual models from natural language supervision,'' in \emph{Proceedings of the 38th International Conference on Machine Learning}, ser. PMLR, vol. 139.\hskip 1em plus 0.5em minus 0.4em\relax PMLR, 2021, pp. 8748--8763.

\bibitem{li2023blip2}
J.~Li, D.~Li, S.~Savarese, and S.~C.~H. Hoi, ``Blip-2: Bootstrapping language-image pre-training with frozen image encoders and large language models,'' in \emph{Proceedings of the 40th International Conference on Machine Learning}, ser. PMLR, vol. 202.\hskip 1em plus 0.5em minus 0.4em\relax Honolulu, HI: PMLR, 2023, pp. 19\,730--19\,742.

\bibitem{hardt2016equality}
M.~Hardt, E.~Price, and N.~Srebro, ``Equality of opportunity in supervised learning,'' vol.~29, 2016.

\bibitem{chouldechova2017fair}
A.~Chouldechova, ``Fair prediction with disparate impact: A study of bias in recidivism prediction instruments,'' \emph{Big Data}, vol.~5, no.~2, pp. 153--163, 2017.

\bibitem{ntoutsi2020bias}
E.~Ntoutsi, P.~Fafalios, U.~Gadiraju, V.~Iosifidis, W.~Nejdl, M.-E. Vidal, S.~Ruggieri, F.~Turini, S.~Papadopoulos, E.~Krasanakis, and et~al., ``Bias in data-driven artificial intelligence systems—an introductory survey,'' \emph{Wiley Interdisciplinary Reviews: Data Mining and Knowledge Discovery}, vol.~10, no.~3, p. e1356, 2020.

\bibitem{ghanbarzadeh2023gender}
S.~Ghanbarzadeh, Y.~Huang, H.~Palangi, R.~Cruz~Moreno, and H.~Khanpour, ``Gender-tuning: Empowering fine-tuning for debiasing pre-trained language models,'' \emph{arXiv preprint arXiv:2307.10522}, 2023.

\bibitem{xie2023parameter}
Z.~Xie and T.~Lukasiewicz, ``An empirical analysis of parameter-efficient methods for debiasing pre-trained language models,'' \emph{arXiv preprint arXiv:2306.04067}, 2023.

\bibitem{xue2024bmft}
Y.~Xue, J.~Yan, R.~Dutt, F.~Haider, J.~Liu, S.~McDonagh, and S.~A. Tsaftaris, ``Bmft: Achieving fairness via bias-based weight masking fine-tuning,'' \emph{arXiv preprint arXiv:2408.06890}, 2024.

\bibitem{jung2024unified}
H.~Jung, T.~Jang, and X.~Wang, ``A unified debiasing approach for vision-language models across modalities and tasks,'' in \emph{Advances in Neural Information Processing Systems}, vol.~37, 2024.

\bibitem{xue2025dfl}
Y.~Xue, J.~Yan, R.~Dutt, F.~Haider, J.~Liu, S.~McDonagh, and S.~A. Tsaftaris, ``Deep fair learning: A unified framework for fine-tuning and debiasing pre-trained models,'' \emph{arXiv preprint arXiv:2504.06470}, 2025.

\bibitem{voigt2017gdpr}
P.~Voigt and A.~Von~dem Bussche, \emph{The EU General Data Protection Regulation (GDPR): A Practical Guide}, 1st~ed.\hskip 1em plus 0.5em minus 0.4em\relax Cham: Springer International Publishing, 2017.

\bibitem{Luo2024FairCLIP}
Y.~Luo, M.~Shi, M.~O. Khan, M.~M. Afzal, H.~Huang, S.~Yuan, Y.~Tian, L.~Song, A.~Kouhana, T.~Elze, Y.~Fang, and M.~Wang, ``{FairCLIP}: Harnessing fairness in vision–language learning,'' in \emph{arXiv preprint arXiv:2403.19949}, 2024.

\bibitem{Hashimoto2018Fairness}
T.~Hashimoto, M.~Srivastava, H.~Namkoong, and P.~Liang, ``Fairness without demographics in repeated loss minimization,'' in \emph{Proceedings of the 35th International Conference on Machine Learning}, 2018, pp. 1929--1938.

\bibitem{chai2022self}
J.~Chai and X.~Wang, ``Self-supervised fair representation learning without demographics,'' \emph{Advances in Neural Information Processing Systems}, vol.~35, pp. 27\,100--27\,113, 2022.

\end{thebibliography}

\end{document}